\documentclass{iopjournal}

\usepackage{amsmath,amssymb,amsfonts}
\usepackage{cite}
\usepackage{graphicx}
\graphicspath{{figures/}}
\usepackage{textcomp}
\usepackage{xcolor}
\usepackage{booktabs}
\usepackage{multirow}
\usepackage{url}
\usepackage{tabularx}
\usepackage{threeparttable}
\usepackage[caption=false,font=footnotesize]{subfig}
\usepackage{wrapfig}
\usepackage{placeins}
\usepackage{pifont}
\usepackage{chngcntr}

\newcommand{\cmark}{\ding{51}}%
\newcommand{\xmark}{\ding{55}}%

\begin{document}

\articletype{Paper}

\title{Energy-Aware Spike Budgeting for Continual Learning in Spiking Neural Networks for Neuromorphic Vision}

\author{Anika Tabassum Meem$^1$\orcid{0009-0002-8444-9923}, Muntasir Hossain Nadid$^1$\orcid{0009-0007-0215-2269}, and Md Zesun Ahmed Mia$^{1,2,*}$\orcid{0009-0004-3509-8455}}

\affil{$^1$ Dept. of Electrical and Electronics Engineering, University of Liberal Arts Bangladesh, Dhaka, Bangladesh\\
$^2$ Dept. of Electrical Engineering, Pennsylvania State University, University Park, USA\\
$^*$ Corresponding author}

\email{anika.tabassum.eee@ulab.edu.bd, muntasir.hossain.eee@ulab.edu.bd, zesun.ahmed@psu.edu}

\begin{abstract}
Neuromorphic vision systems based on spiking neural networks (SNNs) offer ultra-low-power perception for event-based and frame-based cameras, yet catastrophic forgetting remains a critical barrier to deployment in continually evolving environments. Existing continual learning methods, developed primarily for artificial neural networks, seldom jointly optimize accuracy and energy efficiency, with particularly limited exploration on event-based datasets. We propose an energy-aware spike budgeting framework for continual SNN learning that integrates experience replay, learnable leaky integrate-and-fire neuron parameters, and an adaptive spike scheduler to enforce dataset-specific energy constraints during training. Our approach exhibits modality-dependent behavior: on frame-based datasets (MNIST, CIFAR-10), spike budgeting acts as a sparsity-inducing regularizer, improving accuracy while reducing spike rates by up to 47\%; on event-based datasets (DVS-Gesture, N-MNIST, CIFAR-10-DVS), controlled budget relaxation enables accuracy gains up to 17.45 percentage points with minimal computational overhead. Across five benchmarks spanning both modalities, our method demonstrates consistent performance improvements while minimizing dynamic power consumption, advancing the practical viability of continual learning in neuromorphic vision systems.
\end{abstract}


\section{Introduction}
\label{sec:introduction}

Neuromorphic computing has emerged as a paradigm shift in artificial intelligence, drawing inspiration from the energy efficiency and temporal dynamics of the biological brain. Recent research has delved deeper into bio-plausible architectures, ranging from astromorphic transformers \cite{mia2025delving} to bio-inspired sequence processing mechanisms, all aiming to bridge the gap between biological and artificial intelligence \cite{roy2019towards, mia2026rmaat, davies2018loihi}. These systems, particularly those based on spiking neural networks (SNNs), promise ultra-low-power perception by leveraging discrete, event-driven computations compatible with event-based cameras \cite{lichtsteiner2008dvs}. However, deploying these systems in continually changing environments remains challenging due to catastrophic forgetting---the abrupt loss of previously learned knowledge when new tasks are introduced---a central failure mode in continual learning \cite{mccloskey1989catastrophic}. 

The continual learning literature has developed numerous strategies to mitigate forgetting in artificial neural networks. Regularization-based approaches, such as Elastic Weight Consolidation (EWC) \cite{kirkpatrick2017ewc} and Synaptic Intelligence \cite{zenke2017continual}, selectively consolidate important parameters to preserve past knowledge. Architectural methods like Progressive Neural Networks \cite{rusu2016progressive} and PackNet \cite{mallya2018packnet} allocate dedicated capacity for new tasks. Memory-based approaches, including Experience Replay \cite{rolnick2019experience}, Gradient Episodic Memory (GEM) \cite{lopezpaz2017gem}, and online retrieval methods \cite{aljundi2019online, chaudhry2019tiny}, explicitly store and rehearse past examples. While effective for dense, synchronous ANNs, these methods rarely consider the unique constraints and opportunities of event-driven, sparse spiking architectures.

Extending these principles to SNNs introduces additional complexity. Recent work in spiking continual learning has explored Bayesian approaches \cite{skatchkovsky2022bayesian} that leverage probabilistic inference for uncertainty-aware learning, context-aware mechanisms \cite{chen2024contextaware} that adaptively modulate network resources based on task similarity, and on-chip learning frameworks \cite{stewart2020onchip, daram2023neo} that enable real-time adaptation on neuromorphic hardware. Domain-specific applications have also been investigated, including the ICONS '25 semi-supervised neuromorphic cybersecurity pipeline \cite{mia2025neuromorphic_cybersecurity} and temporal sequence processing. However, these methods either assume sparsity is implicitly encouraged by rate coding or focus on architectural heuristics that do not generalize to high-resolution event-based benchmarks, leaving energy-use and accuracy tightly coupled. This assumption breaks down when spike budgets must adapt to diverse input modalities—particularly when contrasting dense Poisson-encoded frames against naturally sparse event streams.

Several prior works improve spiking continual learning by modifying consolidation mechanisms (e.g., Bayesian uncertainty \cite{skatchkovsky2022bayesian}), leveraging task-context similarity \cite{chen2024contextaware}, or introducing neuron-state-dependent plasticity \cite{daram2023neo}. For context, HLOP reports strong retention on permuted-MNIST under a domain-incremental protocol (e.g., ACC 95.15\% with BWT $-1.30\%$ under DSR+HLOP) \cite{xiao2024hebbian}, and TACOS reports mean accuracy of 82.56\% (split-MNIST) and 93.22\% (split-Fashion-MNIST) under task-agnostic domain-incremental learning \cite{soures2024tacos}; see Appendix~\ref{app:prior_work_context} (Table~\ref{tab:prior_work_context}) for a concise protocol-aware comparison. Adaptive pathway reorganization targets performance--energy--memory co-optimization on large-scale CL benchmarks (e.g., CIFAR100/ImageNet splits) \cite{han2023adaptive}, and CoLaNET reports strong continual-learning performance on permuted-MNIST protocols \cite{larionov2025continual}. While these results are promising, they are typically evaluated under different protocols, architectures, and datasets, so direct numerical comparison is not always meaningful. Our work is complementary: we focus on the accuracy--energy trade-off in continual spiking vision across both frame-based and native event-based datasets, and show that explicit spike-budget control can systematically reshape forgetting behavior while keeping activity within neuromorphic sparsity regimes.

To bridge this gap, we propose an energy-aware spike budgeting framework for continual learning in SNNs. We evaluate our method across both frame-based (MNIST, CIFAR-10) and event-based (N-MNIST, CIFAR-10-DVS, DVS-Gesture) datasets. Our approach combines experience replay, learnable neuron dynamics, and an adaptive spike scheduler to enforce dataset-specific energy constraints. On frame-based datasets, our method acts as a sparsity-inducing regularizer, reducing spikes by 47\% on MNIST while improving accuracy by 2.31 percentage points. Conversely, on event-based datasets like DVS-Gesture, relaxing the budget allows for a significant accuracy gain (17.45 percentage points) with only a marginal increase in spike activity, demonstrating the importance of modality-adaptive energy regulation.
Our contributions are summarized as follows:
\begin{itemize}
    \item \textbf{Energy-Aware Continual Learning Framework:} We introduce an adaptive spike budgeting mechanism based on proportional control that dynamically regulates network activity during continual learning, treating energy as a first-class optimization objective alongside accuracy.
    
    \item \textbf{Modality-Dependent Duality Discovery:} We reveal and formalize a fundamental duality in SNN continual learning: spike budgeting acts as sparsity regularization on frame-based datasets (preventing overfitting) while enabling controlled activation increase on event-based datasets (preventing underfitting).
    
    \item \textbf{Cross-Modality Evaluation and Analysis:} We provide a comprehensive study of continual learning spanning five benchmarks across both frame-based and event-based vision, demonstrating consistent improvements in the accuracy-energy trade-off and strong absolute performance (91.93\%) on DVS-Gesture.
\end{itemize}
Our results highlight the importance of energy-aware, modality-adaptive learning, bringing continual SNN learning closer to practical neuromorphic deployment.

\section{Preliminaries}
\label{sec:preliminaries}

This section provides the essential background required to understand our method. We briefly review the leaky integrate-and-fire (LIF) neuron model with surrogate-gradient learning, define the spike-rate energy metric used throughout the paper, and summarize the continual learning evaluation metrics used in our experiments.

\subsection{LIF Neuron Model and Surrogate Gradients}

All networks in this work utilize discrete-time Leaky Integrate-and-Fire (LIF) neurons, simulated via the \texttt{snnTorch} framework \cite{eshraghian2023snntorch}. The membrane potential $U[t]$ of a neuron at time step $t$ evolves according to the following difference equation:

\begin{equation}
U[t+1] = \beta U[t] + I_{\text{in}}[t+1] - S[t] V_{\text{thr}},
\label{eq:lif_update}
\end{equation}
where $\beta \in (0, 1)$ is the membrane potential decay rate, $I_{\text{in}}[t+1]$ is the synaptic input current (weighted sum of incoming spikes), and $V_{\text{thr}}$ is the firing threshold. A spike $S[t] \in \{0, 1\}$ is emitted if the membrane potential exceeds the threshold:

\begin{equation}
S[t] = \Theta(U[t] - V_{\text{thr}}),
\end{equation}
where $\Theta(\cdot)$ is the Heaviside step function. The term $-S[t] V_{\text{thr}}$ in Eq.~\eqref{eq:lif_update} implements a soft reset (reset-by-subtraction) mechanism, which is known to preserve residual potential and mitigate information loss compared to hard resets \cite{han2020rmp}.
To enable training via backpropagation through time (BPTT), we address the non-differentiability of the spike function using the Fast Sigmoid surrogate gradient \cite{zenke2018superspike}, defined as:

\begin{equation}
\frac{\partial S}{\partial U} \approx \frac{1}{(1 + k|U - V_{\text{thr}}|)^2},
\end{equation}
where $k$ is a slope parameter governing the steepness of the surrogate derivative. We use $k = 25$ (the \texttt{snnTorch} default) and favor the Fast Sigmoid because it offers a good balance between gradient stability and task accuracy \cite{zenke2018superspike}. This formulation ensures smooth gradient flow during the backward pass while maintaining binary spiking dynamics in the forward pass.

\subsection{Spike-Rate as an Energy Proxy}

Consistent with neuromorphic hardware literature on Intel Loihi \cite{davies2018loihi}, IBM TrueNorth \cite{akopyan2015truenorth}, and broader neuromorphic computing analyses \cite{roy2019towards}, we treat spike rate as a proxy for energy consumption. For each sample, the spike rate is computed as the average number of spikes per neuron per timestep:

\begin{equation}
\text{SpikeRate} = \frac{1}{N T} 
\sum_{n=1}^{N} \sum_{t=1}^{T} z_{n,t},
\end{equation}
where $N$ is the number of neurons and $T$ is the number of simulation timesteps. Throughout this paper, we report spike rate in percent, i.e., $\text{SpikeRate}(\%) = 100 \times \text{SpikeRate}$. We treat this spike rate as our primary proxy for dynamic energy consumption, as power in neuromorphic hardware is dominated by synaptic operations triggered by active spikes \cite{davies2018loihi}. This metric ignores static leakage power, which is constant across models, focusing instead on the activity-dependent component relevant to algorithm design.
Frame-based datasets naturally induce denser spike activity through Poisson rate coding, while event-based datasets produce inherently ultra-sparse activations.

\subsection{Continual Learning Metrics}

We adopt standard continual-learning metrics following \cite{parisi2019continual}. After training all tasks sequentially, the final average accuracy (ACC) is defined as:

\begin{equation}
\text{ACC} = \frac{1}{K} \sum_{k=1}^{K} A_{K,k},
\end{equation}
where $A_{K,k}$ denotes accuracy on task $k$ after completing all $K$ tasks.
Forgetting is measured as the reduction from each task’s maximal observed accuracy:

\begin{equation}
F = \frac{1}{K-1} \sum_{k=1}^{K-1} 
\left( \max_{j \leq K} A_{j,k} - A_{K,k} \right).
\end{equation}

Backward Transfer (BWT) measures the influence of learning new tasks on the performance of previously learned tasks. It is defined as:

\begin{equation}
\text{BWT} = \frac{1}{K-1} \sum_{k=1}^{K-1} (A_{K,k} - A_{k,k}).
\end{equation}
A positive BWT indicates that learning new tasks has improved the performance on older tasks, while a negative BWT implies catastrophic forgetting.

All models are implemented in PyTorch using snnTorch and trained end-to-end with surrogate-gradient backpropagation. Full architectural and training details are provided in Appendix~\ref{app:implementation}.

\section{Method: Energy-Aware Continual Learning in SNNs}
\label{sec:method}

\subsection{Spiking Network Architectures}
\label{subsec:architectures}
In this work, we employ a family of spiking neural networks (SNNs) across all datasets, ensuring that every model uses leaky integrate-and-fire (LIF) neurons trained with surrogate-gradient backpropagation. No ANN or non-spiking baselines are used; all architectures are fully spiking networks (with either convolutional or fully-connected backbones depending on the dataset).

Our architectures follow a unified design principle adapted to dataset complexity and input resolution. For low-resolution tasks (MNIST, Fashion-MNIST), we use a lightweight fully-connected SNN (Linear--LIF--Linear). For CIFAR-10, we use a VGG-style spiking CNN. For high-resolution event streams (CIFAR-10-DVS, DVS-Gesture), we use a deeper spiking CNN with a compact classifier head to capture fine-grained spatiotemporal features. Detailed layer specifications and tensor shapes are provided in Appendix~\ref{app:implementation}.

\subsection{Energy-Aware Spike Budgeting Framework}
\label{subsec:spike_budgeting_method}
A core contribution of this work is the integration of an energy-aware spike scheduler that dynamically regulates network activity during training. Motivated by the energy constraints of neuromorphic hardware, where power consumption is dominated by synaptic events \cite{davies2018loihi, akopyan2015truenorth}, we introduce a mechanism to enforce a dataset-specific spike budget.

We formulate spike budgeting as a feedback control problem. This mechanism mimics homeostatic plasticity principles \cite{turrigiano1999homeostatic} by dynamically adjusting a regularization coefficient $\lambda_{\text{rate}}$ based on the error between the current rate and a target budget $r_{\text{target}}$:

\begin{equation}
\Delta \lambda_{\text{rate}} = \eta \cdot (r_{\text{spike}} - r_{\text{target}}),
\end{equation}

where $\eta$ is the controller gain. The regularization strength $\lambda_{\text{rate}}$ is updated periodically and clipped to a predefined range $[\lambda_{\min}, \lambda_{\max}]$. This coefficient weights a penalty term in the loss function:

\begin{equation}
\mathcal{L}_{\text{total}} = \mathcal{L}_{\text{task}} + \lambda_{\text{rate}} \cdot (r_{\text{spike}} - r_{\text{target}})^2.
\end{equation}

The squared penalty provides a bidirectional gradient with respect to $r_{\text{spike}}$: when $r_{\text{spike}} > r_{\text{target}}$, $\frac{\partial \mathcal{L}}{\partial r_{\text{spike}}} = 2\lambda_{\text{rate}}(r_{\text{spike}} - r_{\text{target}}) > 0$ suppresses activity; when $r_{\text{spike}} < r_{\text{target}}$, the gradient is negative and would favor increased activity. Crucially, the proportional controller modulates how strongly this term competes with the task loss: if spikes exceed the target, $\lambda_{\text{rate}}$ increases, enforcing sparsity; if spikes fall short, $\lambda_{\text{rate}}$ decreases, effectively relaxing the energy constraint so that $\mathcal{L}_{\text{task}}$ can drive the amount of activity needed for accurate temporal feature extraction.

This mechanism exhibits modality-adaptive behavior. On dense frame-based datasets, where spike rates naturally exceed the target, the controller enforces sparsity. On ultra-sparse event-based datasets, where spike rates fall below the target, the bidirectional penalty gradient encourages increased activity to improve temporal feature extraction. For detailed spike-rate measurement (Appendix~\ref{app:spike_rate_measurement}), controller dynamics (Appendix~\ref{app:budget_loss_controller}), and practical tuning ranges (Appendix~\ref{app:practical_settings}), see Appendix~\ref{app:spike_budgeting}.

\subsection{Learnable Neuron Dynamics}
\label{subsec:learnable_dynamics}

In addition to the spike budgeting framework, we augment the network's representational capacity through learnable LIF neuron parameters. Standard LIF neurons typically use fixed decay rates and thresholds. To enhance the network's ability to capture diverse temporal dynamics---ranging from static frames to fast-changing event streams---we make the membrane decay parameter $\beta$ and the firing threshold $V_{\text{thr}}$ learnable. These are implemented as layer-wise scalars, adding only negligible parameter overhead ($O(L)$ rather than $O(N)$), ensuring the memory footprint remains virtually unchanged. This allows the network to adapt its temporal integration window and sensitivity to the specific statistics of the input data \cite{fang2021incorporating}.

\subsection{Continual Learning Components}
\label{subsec:cl_components}

Having established the energy budgeting mechanism and learnable neuron dynamics, we now describe the continual learning protocol and memory consolidation strategy that complete our framework.

\subsubsection{Class-Incremental Protocol}
\label{subsubsec:class_il}
We adopt a standard class-incremental continual learning (Class-IL) protocol. The dataset is partitioned into a sequence of $K$ disjoint tasks, $\mathcal{T} = \{T_1, T_2, \dots, T_K\}$, where each task contains a unique subset of classes. The model is trained sequentially on each task without access to task identifiers during inference. This setting is particularly challenging as it requires the model to discriminate between all classes seen so far without catastrophic forgetting.

\subsubsection{Experience Replay}
To mitigate forgetting, we employ experience replay, a widely effective strategy in continual learning \cite{rolnick2019experience}. We maintain a fixed-size episodic memory buffer $\mathcal{M}$ that stores a class-balanced subset of samples from previous tasks. During training on task $T_k$, minibatches are constructed by concatenating current-task samples with a random subset drawn from $\mathcal{M}$. This ensures that the optimization objective approximates the joint distribution of all seen tasks, stabilizing the plasticity-stability trade-off.
Replay budgets used across datasets are summarized in Appendix~\ref{app:replay_buffer_budgets} (Table~\ref{tab:replay_budgets}).

\subsection{Ablation Framework}
\label{subsec:ablation_framework}

To systematically evaluate the contributions of each component, we define a hierarchy of experimental configurations that enable controlled ablation studies:

\begin{itemize}
    \item \textbf{C0 (Naive Baseline):} The SNN is trained sequentially on tasks without any replay or regularization. This serves as a lower bound to quantify the severity of catastrophic forgetting.
    \item \textbf{C1 (Replay Baseline):} Adds the experience replay buffer to the naive baseline. This represents the standard approach for mitigating forgetting in SNNs.
    \item \textbf{C2 (Learnable Dynamics):} Augments C1 with learnable LIF parameters ($\beta, V_{\text{thr}}$), isolating the benefit of adaptive temporal integration.
    \item \textbf{C3 (Energy Scheduler):} Augments C1 with the energy-aware spike scheduler, isolating the impact of spike budgeting on accuracy and energy efficiency.
    \item \textbf{C4 (Combined Framework):} Integrates all components: experience replay, learnable dynamics, and the energy-aware scheduler. This represents our proposed holistic approach.
\end{itemize}

\section{Experiments}
\label{sec:experiments}
\subsection{Experimental Setup}
\label{sec:experimental_setup}

We evaluate the proposed framework on five benchmarks spanning frame-based and event-based modalities: MNIST \cite{lecun1998mnist}, N-MNIST \cite{orchard2015converting}, CIFAR-10 \cite{krizhevsky2009learning}, CIFAR-10-DVS \cite{li2017cifar10dvs}, and DVS-Gesture \cite{amir2017gesture}. All experiments follow the class-incremental (Class-IL) protocol, a challenging continual learning setting where models encounter disjoint class subsets sequentially without access to task identifiers at inference \cite{parisi2019continual, zenke2017continual}. Table~\ref{tab:datasets} summarizes the dataset configurations.

\begin{table}[t]
\centering
\caption{Dataset specifications. $T$ denotes timesteps for temporal integration.}
\label{tab:datasets}
%\resizebox{\columnwidth}{!}{%
\begin{tabular}{l c c c c l}
\toprule
\textbf{Dataset} & \textbf{Res.} & \textbf{Classes} & \textbf{Tasks} & \textbf{T} & \textbf{Modality} \\
\midrule
MNIST & $28\times28$ & 10 & 5$\times$2 & 25 & Frame (Poisson) \\
N-MNIST & $34\times34$ & 10 & 5$\times$2 & 50 & DVS (events) \\
CIFAR-10 & $32\times32$ & 10 & 5$\times$2 & 50 & Frame (Poisson) \\
CIFAR-10-DVS & $128\times128$ & 10 & 5$\times$2 & 50 & DVS (events) \\
DVS-Gesture & $128\times128$ & 11 & 4+4+3 & 60 & DVS (events) \\
\bottomrule
\end{tabular}%
%}
\end{table}
\noindent
Tasks denotes the class-incremental schedule (e.g., $5\times2$ indicates five tasks with two classes each, whereas $4+4+3$ covers three tasks with 4, 4, and 3 classes). These timestep budgets reflect the typical temporal extent of each dataset: CIFAR-10-DVS clips comprise short fixed-duration event windows, whereas DVS-Gesture requires $T=60$ to cover full gesture sequences without incurring excessive spike counts.

\textbf{Implementation.} Frame-based datasets are converted to spike trains via Poisson rate encoding \cite{diehl2015snn}, while event-based datasets retain their native DVS representation \cite{lichtsteiner2008dvs, gallego2020eventsurvey}. All models employ LIF-based spiking architectures trained end-to-end with surrogate gradient backpropagation \cite{neftci2019surrogate, wu2018spatio}. The adaptive spike scheduler enforces dataset-specific energy constraints, providing hardware-relevant signals as synaptic operations dominate power in neuromorphic processors \cite{davies2018loihi, akopyan2015truenorth}. Full architectural specifications plus hyperparameters are detailed in Appendix~\ref{app:implementation}.

\begin{figure}[t]
\centering
\includegraphics[width=0.95\linewidth]{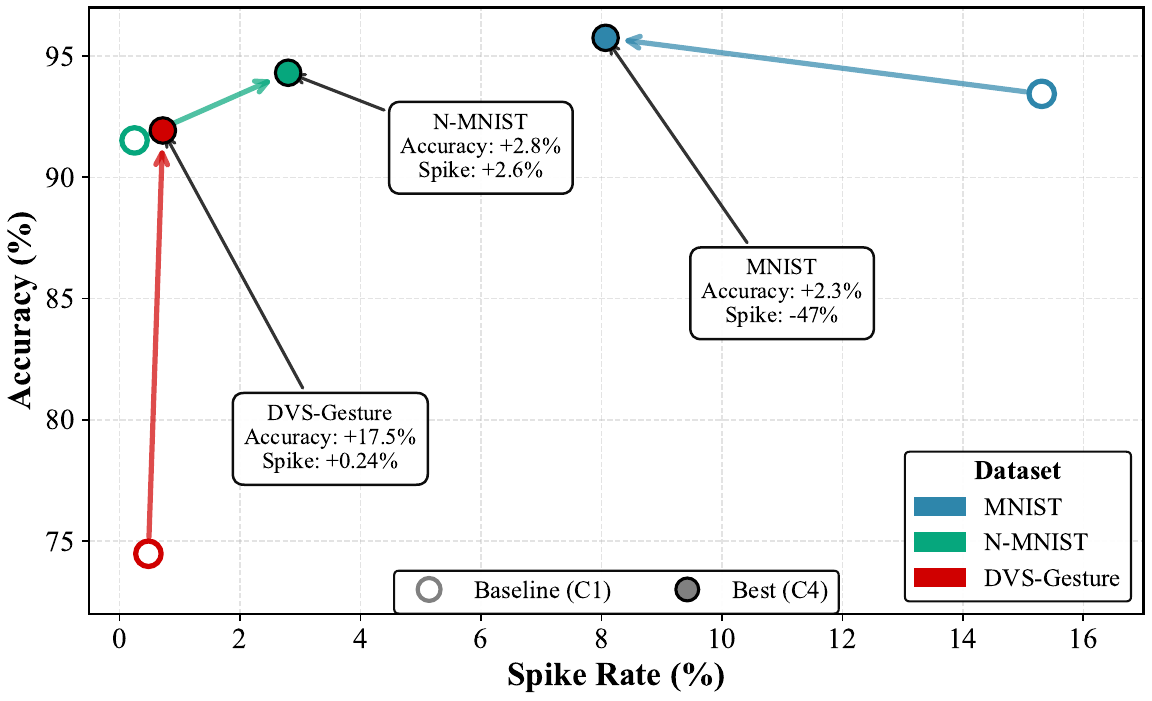}
\caption{Accuracy-energy trade-off across representative datasets. Arrows indicate improvement from C1 (hollow circles) to best configuration (filled circles). MNIST achieves both accuracy improvement and spike reduction. DVS datasets show controlled spike increases that enable substantial accuracy gains while remaining ultra-sparse (below $1\%$ on DVS-Gesture and below $8\%$ across all DVS benchmarks).}
\label{fig:spike_energy}
\end{figure}

\textbf{Evaluation Metrics.} Following standard continual learning methodology \cite{parisi2019continual, lopezpaz2017gem}, we report: (i) average accuracy across all tasks, (ii) forgetting, measured as mean accuracy degradation from peak performance, (iii) backward transfer (BWT), and (iv) spike rate as an energy proxy. Unless stated otherwise, spike rate (\%) is computed during training as the mini-batch average activity (Appendix~\ref{app:spike_rate_measurement}) and then averaged across the training process. All metrics are averaged over multiple random seeds (Appendix~\ref{app:extended_results}). Inference-time spike rates were consistent with training-time averages.

\subsection{Main Results}
\label{subsec:main_results}

Table~\ref{tab:main_results} compares our best configurations against the experience replay baseline (C1) \cite{rolnick2019experience}. Without continual learning (C0), catastrophic forgetting reduces accuracy to $\sim$15--19\% across all datasets (see Appendix~\ref{app:extended_results} for per-seed evidence). Experience replay (C1) establishes a strong baseline, which our energy-aware framework further improves.

\begin{table}[h]
\centering
%\begin{threeparttable}
\caption{Comparison of replay baseline (C1) with best configuration (C4). Bold indicates best performance.}
\label{tab:main_results}
\begin{tabular}{l c c c c c}
\toprule
\textbf{Dataset} & \textbf{Cfg} & \textbf{Acc ($\%$)$^\dagger$} & \textbf{Forgetting (\%)} & \textbf{BWT (\%)} & \textbf{Spike (\%)} \\
\midrule
\multirow{2}{*}{MNIST} 
 & C1 & $93.44\pm1.40$ & 7.17 & -2.45 & 15.31 \\
 & C4 & $\textbf{95.75}\pm0.16$ & \textbf{4.40} & \textbf{-1.65} & \textbf{8.07} \\
\midrule
\multirow{2}{*}{N-MNIST} 
 & C1 & $92.27\pm2.43$ & 6.09 & -8.59 & 0.25 \\
 & C4 & $\textbf{94.07}\pm1.35$ & \textbf{4.06} & \textbf{-5.37} & \textbf{2.70} \\
\midrule
\multirow{2}{*}{CIFAR-10} 
 & C1 & $59.50\pm1.58$ & 33.56 & -60.37 & 37.50 \\
 & C4 & $\textbf{61.26}\pm0.79$ & 33.56 & \textbf{-60.16} & \textbf{30.90} \\
\midrule
\multirow{2}{*}{CIFAR-10-DVS}
 & C1 & $45.52\pm4.24$ & 37.11 & -38.19 & 1.18 \\
 & C4 & $\textbf{49.68}\pm0.93$ & \textbf{29.51} & \textbf{-25.07} & \textbf{7.25} \\
\midrule
\multirow{2}{*}{DVS-Gesture}
 & C1 & $74.48\pm0.74$ & 25.35 & -25.35 & 0.48 \\
 & C4 & $\textbf{91.93}\pm1.85$ & \textbf{5.90} & \textbf{-5.90} & \textbf{0.72} \\
\bottomrule
\end{tabular}

\vspace{2pt}
{\footnotesize $\dagger$ Mean $\pm$ sample standard deviation over the seed subsets detailed in Appendix~\ref{app:extended_results}.}
\end{table}

Results are averaged over multiple independent training trials to ensure statistical robustness. The substantial BWT improvements ($-25.35\%\rightarrow -5.90\%$ for DVS-Gesture) demonstrate the stability of our method across repeated runs.

Before unpacking the per-modality trends, Figure~\ref{fig:spike_energy} summarizes how each optimized configuration shifts relative to its replay baseline in the accuracy--energy plane, making the emerging trade-offs explicit.

\textbf{Modality-Dependent Behavior.} Two distinct patterns emerge across modalities. On frame-based datasets (MNIST, CIFAR-10), spike budgeting acts as implicit regularization, improving both accuracy and energy efficiency. This sparsity-inducing effect is particularly pronounced on MNIST, where the 47\% spike reduction occurs alongside accuracy improvements, suggesting that dense Poisson encoding induces redundant activations that can be pruned without information loss. On event-based datasets (N-MNIST, CIFAR-10-DVS, DVS-Gesture), controlled increases in spike activity yield substantial accuracy gains while maintaining ultra-sparse operation. This aligns with SNN theory \cite{tavanaei2019deep, roy2019towards}: event-driven inputs require sufficient network activation for temporal feature extraction, and overly aggressive sparsity constraints undermine the temporal dynamics necessary for processing asynchronous event streams.

\textbf{Cross-Dataset Magnitude Analysis.} The magnitude of improvement varies significantly across datasets. DVS-Gesture exhibits the largest accuracy gain (17.45 percentage points), several times larger than gains on other datasets. This exceptional improvement stems from three factors: (1) high temporal complexity of gesture sequences requiring adaptive dynamics, (2) relatively low replay baseline (74.48\%) leaving substantial room for improvement, and (3) optimal synergy between learnable LIF parameters and spike budgeting for moderate temporal sequences (T=60). In contrast, N-MNIST shows modest gains (1.80 percentage points) from a higher baseline (92.27\%), with forgetting reduced from 6.09\% to 4.06\%. This suggests that simpler event-based classification tasks approach performance saturation with standard replay mechanisms, though still benefit from reduced forgetting.

\textbf{Configuration Selection.} Across modalities, C4 provides the best accuracy–energy compromise by combining replay, learnable dynamics, and spike budgeting \cite{fang2021incorporating, bellec2018longshort}. CIFAR-10 still benefits primarily from the scheduler term, while the learnable neuron parameters act as a neutral regularizer, illustrating that the unified configuration remains effective even when individual components contribute unevenly. This behavior reinforces the broader design principle: adaptive temporal parameters are most valuable for highly dynamic inputs (DVS streams) but remain compatible with frame-based data when spike budgeting curbs any excess capacity.

\subsection{Ablation Study}
\label{subsec:ablation}

To quantify the contribution of each framework component, we conduct systematic ablation studies. Table~\ref{tab:ablation} presents results on MNIST, illustrating the progression from naive sequential training to our complete method.

\begin{table}[h]
    \centering
    \caption{Ablation study on MNIST (5 tasks $\times$ 2 classes).}
    \label{tab:ablation}
    \begin{tabular}{lccc}
        \toprule
        Configuration & Acc.\ ($\%$)$^\dagger$ & Forgetting (\%) & Spike (\%) \\
        \midrule
        C0 (Naive) & $19.35\pm0.48$ & 99.76 & 15.31 \\
        C1 (Replay) & $93.44\pm1.40$ & 7.17 & 15.31 \\
        C2 (Replay + Learnable) & $94.25\pm0.83$ & 6.16 & 15.31 \\
        C3 (Replay + Scheduler) & $92.40\pm1.75$ & 8.39 & 10.47 \\
        \textbf{C4 (All Components)} & $\textbf{95.75}\pm0.16$ & \textbf{4.40} & \textbf{8.07} \\
        \bottomrule
    \end{tabular}

\vspace{2pt}
{\footnotesize $\dagger$ Mean $\pm$ sample standard deviation over three seeds (Appendix~\ref{app:extended_results}).}
\end{table}

Experience replay \cite{rolnick2019experience} is essential, eliminating catastrophic forgetting \cite{mccloskey1989catastrophic} and establishing a strong 93.44\% baseline. The dramatic improvement from C0 to C1 (74.09 percentage points) confirms that replay-based memory consolidation is non-negotiable for continual SNN learning. Learnable LIF parameters (C2) provide a modest accuracy gain (+0.81 percentage points) without energy cost, suggesting that adaptive temporal integration enhances feature discrimination without inducing additional spiking. This zero-energy cost property makes learnable dynamics an appealing augmentation, as it improves plasticity without compromising the efficiency advantages of SNNs.

The scheduler alone (C3) reduces spikes by 32\% but marginally decreases accuracy ($-$1.04 percentage points), revealing an accuracy-energy trade-off when applied in isolation. However, combining all components (C4) achieves synergistic improvements: highest accuracy (+2.31 percentage points over C1), lowest forgetting ($-$2.77 percentage points over C1), and 47\% spike reduction. This synergy suggests that learnable dynamics partially compensate for the reduced expressiveness imposed by spike constraints, while the scheduler's regularization effect mitigates potential overfitting from increased parameter flexibility. The non-additive nature of these improvements underscores the importance of holistic co-design in neuromorphic systems. Full ablation results across all datasets are in Appendix~\ref{app:extended_results}.

\subsection{Analysis by Modality}
\label{subsec:analysis}

\begin{figure}[t]
    \centering
    \includegraphics[width=0.95\linewidth]{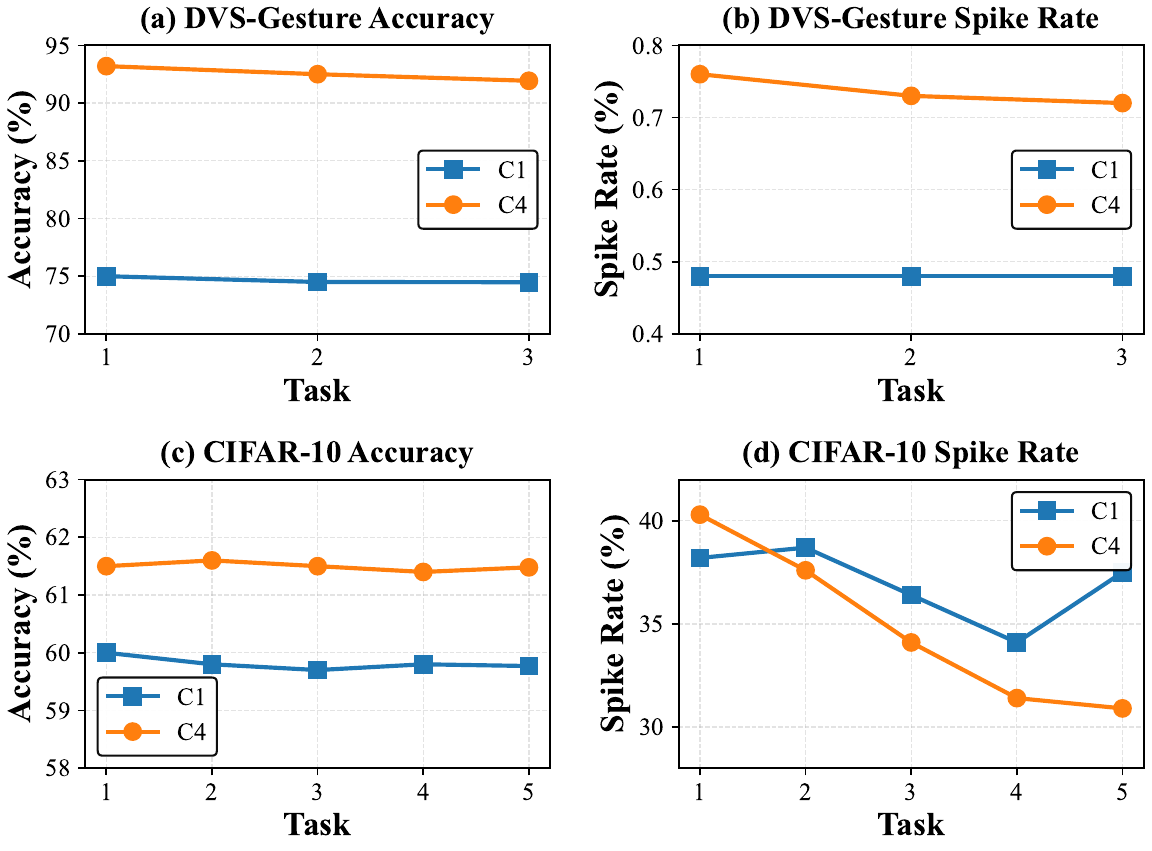}
\caption{Task-wise analysis for representative datasets. (a,b) DVS-Gesture: C4 maintains ultra-low spike rates below 1\% while improving accuracy. (c,d) CIFAR-10: C4 progressively reduces spike rates across tasks.}
    \label{fig:task_analysis}
\end{figure}

\textbf{Event-based datasets.} DVS sensors produce inherently sparse, asynchronous event streams \cite{gallego2020eventsurvey}. On DVS-Gesture \cite{amir2017gesture}, C4 achieves dramatic accuracy gains with minimal spike increase (Fig.~\ref{fig:task_analysis}(a,b)). Task-wise analysis reveals that C4 maintains consistent improvements across all gesture groups, with particularly strong performance on the final task (Task 3), which exhibits the most challenging 3-class discrimination problem. The spike rate remains below 1\% throughout sequential learning, demonstrating that the controlled activation increase (0.48\% $\rightarrow$ 0.72\%) operates well within the ultra-sparse regime characteristic of neuromorphic efficiency \cite{roy2019towards}.

Notably, the absolute spike increase (+0.24 percentage points) is minimal compared to frame-based spike reductions ($-$7.24 percentage points on MNIST), indicating that the scheduler adapts its behavior to input statistics rather than enforcing uniform sparsity constraints. This adaptive property is critical for DVS datasets, where excessive sparsity enforcement can eliminate the temporal information encoded in event timing. The bidirectional gradient mechanism naturally discovers this balance: when spike rates fall below the target, the penalty gradient weakens, allowing the classification loss to drive learning while maintaining hardware-compatible operation.

\textbf{Frame-based datasets.} Poisson-encoded frame data naturally induces denser spike activity. On CIFAR-10, C4 yields a 1.76 percentage-point accuracy improvement and 17.6\% spike reduction reported in Table~\ref{tab:main_results}, with the scheduler component supplying most of the gain and the learnable parameters acting as a neutral regularizer (Fig.~\ref{fig:task_analysis}(c,d)). Task-wise analysis shows that C4 achieves progressive spike reduction across sequential tasks, with particularly strong compression on later tasks (Tasks 4-5), where the replay buffer provides richer supervision. This progressive improvement suggests that the scheduler learns effective regularization as the task distribution stabilizes with experience.

The 17.6\% spike reduction on CIFAR-10 is less pronounced than MNIST's 47\% reduction, reflecting the inherently higher complexity of 32$\times$32 RGB natural images compared to 28$\times$28 grayscale digits. Despite this, the spike rate remains substantially lower than typical dense ANNs would induce via rate encoding, confirming that spike budgeting successfully exploits redundancy in Poisson-encoded representations. Importantly, accuracy improves alongside sparsity, contradicting the intuitive expectation of an accuracy-energy trade-off and validating our hypothesis that redundant spikes can impair generalization through effective overfitting.

\subsection{Energy-Accuracy Trade-offs}
\label{subsec:energy_accuracy}

Figure~\ref{fig:spike_energy} visualizes the accuracy-energy landscape across representative datasets, revealing fundamental differences in optimization trajectories. On MNIST, the improvement vector from C1 to C4 points toward the upper-left quadrant---simultaneous accuracy improvement and energy reduction---a rare and desirable outcome in machine learning. This Pareto improvement demonstrates that spike budgeting on frame-based data does not merely trade accuracy for efficiency but actually enhances both objectives through implicit regularization.

DVS datasets exhibit a different trajectory: improvements vector toward the upper-right quadrant, trading small-to-moderate absolute spike-rate increases for substantial accuracy gains. However, the magnitude of these trade-offs varies dramatically. DVS-Gesture achieves a highly favorable 72:1 accuracy-to-spike ratio (17.45 percentage-point accuracy gain / 0.24 percentage-point spike increase), while CIFAR-10-DVS shows a less favorable 0.68:1 ratio (4.16 percentage-point accuracy gain / 6.07 percentage-point spike increase). This disparity reflects the temporal complexity of gesture recognition versus object classification: gesture sequences (T=60) benefit more from increased representational capacity than shorter CIFAR-10-DVS event windows (T=50).

\textbf{Hardware implications.} Neuromorphic processors such as Loihi \cite{davies2018loihi} and TrueNorth \cite{akopyan2015truenorth} exhibit energy consumption dominated by synaptic operations (AC events). Our 47\% relative spike reduction on MNIST (15.31\% $\rightarrow$ 8.07\%) therefore implies a commensurate reduction in activity-driven synaptic events, all else equal. On DVS-Gesture, despite the +0.24 percentage-point spike increase, absolute spike rates remain below 1\%, maintaining compatibility with neuromorphic hardware specifications where event rates $<10\%$ are considered efficient \cite{roy2019towards}. These results demonstrate that energy-aware training during continual learning preserves deployment viability on resource-constrained neuromorphic platforms.

\section{Discussion}
\label{sec:discussion}

The empirical study shows that energy-aware spike budgeting is not merely a regularizer but a control primitive that reshapes the plasticity--stability trade-off in modality-dependent ways. On frame-based datasets, the controller consistently drives C4 toward lower spike regimes than replay-only baselines while preserving or improving accuracy, indicating that redundant Poisson activations can be trimmed without hurting separability. Conversely, event-based datasets benefit from carefully relaxing the budget: even modest increases in spike rate (0.48\% $\rightarrow$ 0.72\% on DVS-Gesture) supply the temporal support required for long gestures while keeping absolute activity well inside neuromorphic sparsity envelopes. This dual behavior validates our claim that energy should be treated as a first-class optimization signal rather than a passive byproduct of SNN sparsity.

The ablations (Table~\ref{tab:ablation}) clarify why combining replay, learnable neuron dynamics, and the scheduler is essential. Replay alone removes catastrophic forgetting but leaves excess energy on frame data; the scheduler alone enforces budgets yet suffers accuracy loss; learnable dynamics introduce temporal flexibility but require an energy-aware constraint to remain stable. C4 leverages all three: replay supplies representative gradients, learnable $\beta/V_{\text{thr}}$ tune per-dataset time constants, and the controller enforces hardware-aware activity bounds. This decomposition suggests a design rule for future systems—pair any capacity-expanding mechanism (e.g., adaptive neurons or dendritic compartments) with an explicit energy controller to keep continual learning trajectories within feasible regions.

From a deployment standpoint, the learned controller resembles a software-defined activity constraint for neuromorphic hardware. Because the budget is enforced during training, inference-time activity remains predictable without extra calibration, easing mapping onto Loihi- or TrueNorth-like substrates where synaptic operations dominate power cost \cite{davies2018loihi, akopyan2015truenorth}. The ability to dial spike targets per modality further enables heterogeneous sensor suites: a single policy can throttle Poisson-coded cameras aggressively while granting additional headroom to asynchronous event sensors, simplifying firmware integration in mixed-vision platforms.

Despite these advantages, several limitations remain. Replay buffers still add non-trivial memory overhead, which may be challenging for on-chip deployment, and our controller currently relies on hand-tuned gains ($\eta$, $\lambda_{\max}$) per modality. Finally, we focus on class-incremental protocols with balanced task boundaries; task-free or open-world settings may require adaptive targets that respond to concept drift. Addressing these issues—through automatic gain tuning, generative memory surrogates, or adaptive spike targets informed by streaming statistics—remains fertile ground for future neuromorphic continual learning research.

\section{Conclusion}
\label{sec:conclusion}

We presented an energy-aware continual learning framework for spiking neural networks that combines replay, learnable LIF parameters, and a proportional spike scheduler. Across five benchmarks spanning frame-based and event-based vision, this unified approach consistently improved accuracy while either reducing or strategically increasing spike activity, demonstrating that energy budgets can serve as an explicit control signal rather than an afterthought. The modality-dependent behavior—sparsity regularization for dense Poisson inputs versus controlled activation boosts for ultra-sparse DVS streams—highlights the necessity of treating neuromorphic energy constraints as data-aware design parameters.

Looking ahead, we plan to deploy the controller-trained models on neuromorphic hardware to verify real power savings, investigate automatic tuning of budget targets for task-free or open-world streams, and pair the scheduler with lightweight generative replay to reduce memory footprints. These steps will further close the gap between algorithmic continual learning advances and practical neuromorphic perception systems.

%\section*{Acknowledgment}

\providecommand{\newblock}{}

\clearpage

\appendix

% Restart figure, table, and equation numbering for appendix
\setcounter{table}{0}
\setcounter{figure}{0}
\setcounter{equation}{0}
\makeatletter
\renewcommand{\thetable}{\thesection.\arabic{table}}
\renewcommand{\thefigure}{\thesection.\arabic{figure}}
\renewcommand{\theequation}{\thesection.\arabic{equation}}
\@addtoreset{table}{section}
\@addtoreset{figure}{section}
\@addtoreset{equation}{section}
\makeatother

\section{Energy-Aware Spike Budgeting}
\label{app:spike_budgeting}

Configurations C3 and C4 share the same controller but express different ablations of our framework. This appendix documents the measurement strategy, control law, and operating regimes that enable the scheduler to remain stable across modalities.

\subsection{Spike-Rate Measurement}
\label{app:spike_rate_measurement}

We monitor mini-batch activity via
\begin{equation}
    r_{\mathrm{spike}} = \frac{N_{\mathrm{spk}}}{N T B},
\end{equation}
where $N_{\mathrm{spk}}$ counts spikes, $N$ is the number of monitored neurons, $T$ is the integration window, and $B$ is the batch size. We report $100 \times r_{\mathrm{spike}}$ as the spike rate (\%) in all tables and figures. This proxy matches the energy metric in Section~\ref{sec:preliminaries} and is computed once per optimizer step so that controller updates keep pace with weight updates.

\subsection{Budget Loss and Controller}
\label{app:budget_loss_controller}

The total loss for configurations C3/C4 is
\begin{equation}
    \mathcal{L}_{\mathrm{total}} = \mathcal{L}_{\mathrm{task}} + \lambda_{\mathrm{rate}} (r_{\mathrm{spike}} - r_{\mathrm{target}})^2,
\end{equation}
with a clipped proportional controller
\begin{equation}
    \lambda_{\mathrm{rate}} \leftarrow \mathrm{clip}\big( \lambda_{\mathrm{rate}} + \eta (r_{\mathrm{spike}} - r_{\mathrm{target}}), [\lambda_{\min}, \lambda_{\max}] \big).
\end{equation}
Whenever the network overshoots the budget, $\lambda_{\mathrm{rate}}$ increases and the penalty term suppresses spikes; once activity drops below the target, the controller gives back capacity so that the cross-entropy term dominates. In practice, we update $\lambda_{\mathrm{rate}}$ after each optimizer step and cache the resulting value so that the next batch starts from the most recent state.

\subsection{Dataset-Specific Behavior}
\label{app:dataset_specific_behavior}

\textbf{Frame-based datasets.} C1 produces $15.31\%$ spikes on MNIST; the scheduler steers C4 to $8.07\%$ spikes (47\% reduction) while improving accuracy by 2.31 percentage points. The same trend holds for CIFAR-10, where the scheduler acts primarily as a regularizer.

\textbf{Event-based datasets.} DVS-Gesture begins at $0.48\%$ spikes. Allowing a controlled increase to $0.72\%$ unlocks 17.45 percentage points of accuracy while keeping absolute activity below 1\%. CIFAR-10-DVS similarly trades a modest spike increase for 4.16 percentage points of accuracy, albeit with a less favorable accuracy-to-spike ratio because of the shorter temporal windows (T=50).

\subsection{Practical Settings}
\label{app:practical_settings}

We use $r_{\mathrm{target}} = 5$--$12\%$ with $\eta = 0.1$--$0.3$ for frame-based datasets, and $r_{\mathrm{target}} = 0.5$--$3\%$ for DVS datasets, clipping $\lambda_{\mathrm{rate}}$ to $[0, 5]$. Lower targets can silence useful neurons, whereas higher values erode the energy advantages. The scheduler window (5 batches) strikes a balance between responsiveness and stability; shorter windows react too aggressively on CIFAR-10-DVS.

\subsection{Key Takeaways}
\label{app:key_takeaways}

\begin{itemize}
    \item The same control law operates across modalities because the spike proxy normalizes by both hidden units and timesteps, enabling a shared $r_{\mathrm{target}}$ interpretation.
    \item Replay keeps the controller honest: batches always mix old and new samples, so the measured spike rate reflects the deployed distribution rather than a single-task snapshot.
    \item Stability depends on two hyperparameters: the proportional gain $\eta$ and the clipping ceiling $\lambda_{\max}$. We calibrate them once per modality (frame vs.\ event) and then reuse those values for all later experiments, including ablations.
\end{itemize}
\section{Implementation and Training Details}
\label{app:implementation}

Reproducing the study requires four ingredients: how tasks are constructed, which architectures pair with each modality, which optimizer settings unlock stable training, and how the replay buffer is provisioned. The following subsections cover those items in that order so the experimental pipeline can be mirrored end-to-end.

\subsection{Dataset and Task Construction}
\label{app:dataset_task_construction}
Each dataset follows the class-incremental (Class-IL) protocol described in Section~\ref{subsubsec:class_il}: five sequential two-class tasks for MNIST, CIFAR-10, and N-MNIST; five event-based tasks for CIFAR-10-DVS; and a three-stage (4/4/3 gesture) split for DVS-Gesture. Frame-based benchmarks use Poisson rate encoding with $T=25$ (MNIST) or $T=50$ (CIFAR-10). Fashion-MNIST follows the same protocol as MNIST and is included for supplementary validation in Appendix~\ref{app:extended_results} (Table~\ref{tab:seed_fmnist}). Event-based datasets retain their native polarity streams, using $T=50$ for N-MNIST and CIFAR-10-DVS and $T=60$ for the longer DVS-Gesture sequences. The budgeting targets in Appendix~\ref{app:spike_budgeting} are chosen to match these inherent time scales. Under this protocol, the network never revisits earlier tasks during training; by the time CIFAR-10 reaches the fifth task it must already distinguish all ten classes without task labels. The higher timesteps on DVS-Gesture (60 vs.\ 50) stem from the longer gesture duration, which would otherwise be truncated and lead to artificial forgetting unrelated to model capacity.

\subsection{Model Configurations}
\label{app:model_configurations}
The continual-learning variants C0--C4 combine replay, learnable neuron dynamics, and spike budgeting as summarized in Table~\ref{tab:config_mapping}. The architectures themselves follow a consistent motif: lightweight fully-connected SNNs for 28$\times$28 inputs, VGG-style networks for CIFAR-10, and deeper four-block ConvSNNs for the high-resolution DVS datasets. Tables~\ref{tab:mnist-arch}--\ref{tab:dvs-arch} list the layer layouts used throughout the paper.

\begin{table}[t]
\centering
\caption{Component mapping for configurations C0--C4.}
\label{tab:config_mapping}
\begin{tabular}{lccc}
\toprule
Config & Replay & Learnable LIF & Energy Scheduler \\
\midrule
C0 & \xmark & \xmark & \xmark \\
C1 & \cmark & \xmark & \xmark \\
C2 & \cmark & \cmark & \xmark \\
C3 & \cmark & \xmark & \cmark \\
C4 & \cmark & \cmark & \cmark \\
\bottomrule
\end{tabular}
\end{table}

\noindent C0 therefore quantifies catastrophic forgetting, C1 injects replay, C2 adds learnable LIF parameters, C3 activates the spike scheduler, and C4 inherits every mechanism. This explicit mapping becomes important when interpreting later sections: any delta between rows in Table~\ref{tab:main_results} can be traced back to toggling a single column here.

\begin{table}[h]
\centering
\caption{Fully-connected spiking architecture for MNIST and Fashion-MNIST.}
\label{tab:mnist-arch}
\begin{tabular}{llll}
\toprule
Layer & Parameters & Output Size & LIF \\
\midrule
Linear & 784 $\rightarrow$ 128 & 128 & -- \\
LIF & $\beta, V_{\text{thr}}$ learnable (C2/C4) & 128 & \cmark \\
Linear & 128 $\rightarrow$ 10 & 10 & -- \\
\bottomrule
\end{tabular}
\end{table}

\begin{table}[h]
\centering
\caption{VGG-5 Spiking CNN architecture for CIFAR-10.}
\label{tab:cifar-arch}
\begin{tabular}{llll}
\toprule
Layer & Parameters & Output Size & LIF \\
\midrule
Conv2d + BN & 3$\rightarrow$64, k=3, pad=1 & $32\times 32$ & -- \\
Conv2d + BN & 64$\rightarrow$64, k=3, pad=1 & $32\times 32$ & -- \\
MaxPool2d & k=2 & $16\times 16$ & -- \\
LIF & $\beta$, $V_{\text{thr}}$ & $16\times 16$ & \cmark \\
Conv2d + BN & 64$\rightarrow$128, k=3, pad=1 & $16\times 16$ & -- \\
Conv2d + BN & 128$\rightarrow$128, k=3, pad=1 & $16\times 16$ & -- \\
MaxPool2d & k=2 & $8\times 8$ & -- \\
LIF & $\beta$, $V_{\text{thr}}$ & $8\times 8$ & \cmark \\
Conv2d + BN & 128$\rightarrow$256, k=3, pad=1 & $8\times 8$ & -- \\
MaxPool2d & k=2 & $4\times 4$ & -- \\
LIF & $\beta$, $V_{\text{thr}}$ & $4\times 4$ & \cmark \\
Flatten & -- & 4096 & -- \\
Linear & 4096$\rightarrow$10 & 10 & -- \\
\bottomrule
\end{tabular}
\end{table}

\begin{table}[h]
\centering
\caption{Spiking CNN architecture for CIFAR-10-DVS and DVS-Gesture.}
\label{tab:dvs-arch}
\begin{tabular}{llll}
\toprule
Layer & Parameters & Output Size & LIF? \\
\midrule
Conv2d & 2$\rightarrow$32, k=3, pad=1 & $128\times 128$ & -- \\
MaxPool2d & k=2 & $64\times 64$ & -- \\
LIF & $\beta$, $V_{\text{thr}}$ & $64\times 64$ & \cmark \\
Conv2d & 32$\rightarrow$64, k=3, pad=1 & $64\times 64$ & -- \\
MaxPool2d & k=2 & $32\times 32$ & -- \\
LIF & $\beta$, $V_{\text{thr}}$ & $32\times 32$ & \cmark \\
Conv2d & 64$\rightarrow$128, k=3, pad=1 & $32\times 32$ & -- \\
MaxPool2d & k=2 & $16\times 16$ & -- \\
LIF & $\beta$, $V_{\text{thr}}$ & $16\times 16$ & \cmark \\
Conv2d & 128$\rightarrow$256, k=3, pad=1 & $16\times 16$ & -- \\
MaxPool2d & k=2 & $8\times 8$ & -- \\
LIF & $\beta$, $V_{\text{thr}}$ & $8\times 8$ & \cmark \\
Flatten & -- & 16384 & -- \\
Linear & 16384$\rightarrow C$ & $C$ & -- \\
\bottomrule
\end{tabular}
\end{table}

\noindent Listing the architectures per modality prevents ambiguity about channel counts, LIF placements, and flatten dimensions---all of which influence spike statistics. For instance, the DVS model's flatten layer at 16{,}384 features is the main contributor to energy use; the scheduler must therefore regulate activity at that interface, whereas CIFAR-10 relies mainly on convolutional sparsity.

\subsection{Optimization and Training Regime}
\label{app:optimization_training_regime}
All models are trained with Adam. Frame-based datasets use a learning rate of $10^{-3}$, batch size $64$, and five epochs per task; CIFAR-10 uses the same rate but ten epochs with cosine annealing. Event-based datasets use batch size $16$; N-MNIST trains for forty epochs, CIFAR-10-DVS for thirty, and DVS-Gesture for sixty epochs with $3\times10^{-3}$ learning rate and a Reduce-on-Plateau scheduler. Gradient clipping (max norm $= 1.0$) stabilizes the deeper ConvSNNs, and each configuration completes within a few GPU-hours on a single modern accelerator, including replay sampling and controller updates. Longer training schedules were evaluated for DVS-Gesture but provided diminishing returns once the scheduler converged; rather than add a fourth control knob, we fix these hyperparameters for all seeds so that variability stems solely from data order and stochastic updates.
\subsection{Replay Buffer Budgets}
\label{app:replay_buffer_budgets}
The episodic memory $\mathcal{M}$ that powers experience replay (Section~\ref{subsec:cl_components}) is allocated per dataset before training and then partitioned uniformly across classes by the replay buffer. Table~\ref{tab:replay_budgets} enumerates the capacities used for all reported results; the per-class slot count equals $\lfloor |\mathcal{M}| / C \rfloor$ for $C$ classes. For reference, a buffer of 10{,}000 samples corresponds to 20\% of the CIFAR-10 training set (50{,}000 images). Frame-based datasets need larger buffers to counter intra-class variety, whereas DVS-Gesture prioritizes balanced coverage of its 11 gestures while keeping memory footprint modest. A systematic sweep over smaller buffers is left to future work; in this paper we keep the replay budget fixed per dataset to isolate the effect of spike budgeting and learnable dynamics.
\begin{table}[b]
\centering
\caption{Replay memory allocation per dataset.}
\label{tab:replay_budgets}
\begin{tabular}{lccc}
\toprule
\textbf{Dataset} & \textbf{Classes} & \textbf{Buffer Size} & \textbf{Slots / Class} \\
\midrule
MNIST / Fashion-MNIST & 10 & 2000 & 200 \\
N-MNIST & 10 & 2000 & 200 \\
CIFAR-10 & 10 & 10000 & 1000 \\
CIFAR-10-DVS & 10 & 10000 & 1000 \\
DVS-Gesture & 11 & 440 & 40 \\
\bottomrule
\end{tabular}
\end{table}
\section{Extended Results and Ablations}
\label{app:extended_results}

The quantitative evidence that feeds Table~\ref{tab:main_results} is unpacked here via seed-level tables. Each table lists accuracy, forgetting, spike rate, and---where available---BWT so readers can trace how the aggregated numbers were formed.

\subsection{Seed-Level Tables}
\label{app:seed_level_tables}

We list every seed considered per dataset. The tables surface both accuracy/spike trends and BWT statistics where logs provide per-task information, illustrating the variability inherent to deep SNN training and the variability encountered on event-based benchmarks.

%---------------- DVS-Gesture ----------------
\subsubsection{DVS-Gesture}
\label{app:seed_dvs_gesture}

Table~\ref{tab:seed_dvs_gesture} reports per-seed DVS-Gesture metrics under the replay baseline (C1) and the full framework (C4).

\begin{table}[h]
    \centering
    \caption{DVS-Gesture per-seed results.}
    \label{tab:seed_dvs_gesture}
    \begin{tabular}{lcccc}
        \toprule
        Seed & Config & Final Acc.\ (\%) & Forgetting (\%) & Spike Rate (\%) \\
        \midrule
        42 & C1 & 75.00 & 25.87 & 0.48 \\
        42 & C4 & 93.23 & 5.94  & 0.76 \\
        44 & C1 & 73.96 & 24.83 & 0.48 \\
        44 & C4 & 90.62 & 5.86  & 0.67 \\
        \bottomrule
    \end{tabular}
\end{table}

\noindent DVS-Gesture therefore illustrates the regime where relaxing the spike budget is essential: the shift from the replay-only baseline to C4 yields roughly 17--18 percentage points of accuracy while absolute spike rates remain below 1\%, so the energy cost of the improved controller is negligible.

\FloatBarrier

%---------------- CIFAR-10-DVS ----------------
\subsubsection{CIFAR-10-DVS}
\label{app:seed_cifar10dvs}

Table~\ref{tab:seed_cifar10dvs} summarizes performance on CIFAR-10-DVS. The variability across seeds highlights the challenge of training on this dataset; however, C4 consistently reduces forgetting in successful runs.

\begin{table}[h]
    \centering
    \caption{CIFAR-10-DVS per-seed results.}
    \label{tab:seed_cifar10dvs}
    \begingroup
    \setlength{\tabcolsep}{4.5pt}
    \begin{tabular}{lccccc}
        \toprule
        & \multicolumn{2}{c}{Final Acc.\ (\%)} & \multicolumn{2}{c}{Forgetting (\%)} & Spike Rate (\%) \\
        \cmidrule(lr){2-3}\cmidrule(lr){4-5}
        Seed & C1 & C4 & C1 & C4 & C4 \\
        \midrule
        42 & 42.53 & 49.03 & 42.95 & 29.83 & 6.51 \\
        44 & 46.36 & 45.57 & 34.82 & 31.92 & 6.82 \\
        45 & 48.52 & 50.34 & 31.27 & 29.19 & 8.00 \\
        \bottomrule
    \end{tabular}
    \endgroup
\end{table}

\noindent Across CIFAR-10-DVS seeds, C4 narrows forgetting by roughly 3--13 percentage points whenever training remains stable. This table provides the granular evidence behind the average BWT improvement reported in the main text.

\FloatBarrier

%---------------- N-MNIST ----------------
\subsubsection{N-MNIST}
\label{app:seed_nmnist}

Table~\ref{tab:seed_nmnist} shows final accuracy and spike rate on N-MNIST for both C1 and C4.

\begin{table}[h]
    \centering
    \caption{N-MNIST per-seed results for C1 and C4.}
    \label{tab:seed_nmnist}
    \begingroup
    \setlength{\tabcolsep}{5pt}
    \begin{tabular}{lccc}
        \toprule
        Seed & Config & Final Acc.\ (\%) & Spike Rate (\%) \\
        \midrule
        42 & C1 & 93.59 & 0.25 \\
        42 & C4 & 95.60 & 2.74 \\
        44 & C1 & 89.46 & 0.25 \\
        44 & C4 & 93.03 & 2.85 \\
        46 & C1 & 93.76 & 0.25 \\
        46 & C4 & 93.58 & 2.51 \\
        \bottomrule
    \end{tabular}
    \endgroup
\end{table}

\noindent N-MNIST remains largely spike-limited: all C1 seeds operate at $0.25\%$ spikes, whereas C4 permits a small activity increase that converts to 2--4 percentage points of accuracy on the more stable runs. The variability shown here explains the larger standard deviations listed in Table~\ref{tab:main_results}.

\FloatBarrier

%---------------- MNIST ----------------
\subsubsection{MNIST}
\label{app:seed_mnist}

Table~\ref{tab:seed_mnist} details the results for MNIST across all configurations.

\begin{table}[h]
    \centering
    \caption{MNIST per-seed results for C0--C4.}
    \label{tab:seed_mnist}
    \begingroup
    \setlength{\tabcolsep}{4.5pt}
    \begin{tabular}{lcccc}
        \toprule
        Seed & Config & Final Acc.\ (\%) & Forgetting (\%) & Spike Rate (\%) \\
        \midrule
        42 & C0 & 19.84 & 99.82 & 15.31 \\
        43 & C0 & 19.33 & 99.60 & 15.31 \\
        44 & C0 & 18.89 & 99.86 & 15.31 \\
        \midrule
        42 & C1 & 94.93 & 5.54 & 15.31 \\
        43 & C1 & 93.24 & 7.14 & 15.31 \\
        44 & C1 & 92.16 & 8.82 & 15.31 \\
        \midrule
        42 & C2 & 92.22 & 8.71 & 15.31 \\
        43 & C2 & 95.77 & 4.11 & 15.31 \\
        44 & C2 & 94.76 & 5.65 & 15.31 \\
        \midrule
        42 & C3 & 93.07 & 7.87 & 10.87 \\
        43 & C3 & 90.41 & 10.69 & 10.31 \\
        44 & C3 & 93.71 & 6.62 & 10.24 \\
        \midrule
        42 & C4 & 95.59 & 4.70 & 7.64 \\
        43 & C4 & 95.75 & 4.23 & 7.98 \\
        45 & C4 & 95.91 & 4.27 & 8.59 \\
        \bottomrule
    \end{tabular}
    \endgroup
\end{table}

\noindent The MNIST table exposes how each component contributes per seed: replay (C1) already reaches $>93\%$ accuracy, learnable dynamics (C2) maintain the same spike rate while slightly improving fidelity, and the scheduler (C3) alone enforces sparsity. Combining them (C4) yields the consistent $95.7$--$95.9\%$ plateau highlighted in the main paper.

\FloatBarrier

%---------------- CIFAR-10 (frame) ----------------
\subsubsection{CIFAR-10 (Frame-Based)}
\label{app:seed_cifar10} 

Table~\ref{tab:seed_cifar10} shows seed-level results for CIFAR-10 under the full C4 configuration. The scheduler remains the dominant contributor on this benchmark, so the aggregated statistics match the improvements highlighted in the main table.

\begin{table}[h]
    \centering
\caption{CIFAR-10 per-seed results for C1 and C4.}
    \label{tab:seed_cifar10}
    \begingroup
    \small
    \setlength{\tabcolsep}{4.5pt}
    \begin{tabular}{lccccc}
        \toprule
        & \multicolumn{2}{c}{Final Acc.\ (\%)} & \multicolumn{2}{c}{Forgetting (\%)} & Spike Rate (\%) \\
        \cmidrule(lr){2-3}\cmidrule(lr){4-5}
        Seed & C1 & C4 & C1 & C4 & C4 \\
        \midrule
        42 & 60.95 & 60.35 & 32.49 & 32.49 & 30.90 \\
        43 & 57.81 & 61.72 & 33.15 & 33.15 & 30.90 \\
        44 & 59.74 & 61.72 & 35.05 & 35.05 & 30.90 \\
        \bottomrule
    \end{tabular}
    \endgroup
\end{table}

\noindent On CIFAR-10, the scheduler reduces average spike rate (Table~\ref{tab:main_results}) while maintaining or improving accuracy relative to replay-only training. In these runs, forgetting remains comparable between C1 and C4, suggesting that energy control mainly improves generalization/efficiency without destabilizing replay-based consolidation.

\FloatBarrier

%---------------- Fashion-MNIST (Supplementary) ----------------
\subsubsection{Fashion-MNIST (Supplementary Validation)}
\label{app:seed_fmnist}

Fashion-MNIST serves as an additional frame-based benchmark to validate the energy-efficiency findings observed on MNIST. Table~\ref{tab:seed_fmnist} details the results. While accuracy improvements are modest compared to DVS datasets, the energy savings (up to 63\% spike reduction in C4) are substantial even on this supplementary split, reinforcing the sparsity-inducing regularization effect on frame-based vision tasks.

\begin{table}[t]
    \centering
    \caption{Fashion-MNIST per-seed results for C0--C4.}
    \label{tab:seed_fmnist}
    \begingroup
    \small
    \setlength{\tabcolsep}{3.5pt}
    \begin{tabular}{lcccc}
        \toprule
        Seed & Config & Final Acc.\ (\%) & Forgetting (\%) & Spike Rate (\%) \\
        \midrule
        42 & C0 & 19.87 & 98.95 & 32.74 \\
        43 & C0 & 19.88 & 98.96 & 32.74 \\
        44 & C0 & 19.96 & 99.05 & 32.74 \\
        \midrule
        42 & C1 & 80.04 & 21.03 & 32.74 \\
        44 & C1 & 79.48 & 21.84 & 32.74 \\
        45 & C1 & 79.58 & 21.28 & 32.74 \\
 
        \midrule
        42 & C2 & 78.17 & 23.64 & 32.74 \\
        43 & C2 & 79.80 & 21.39 & 32.74 \\
        44 & C2 & 79.90 & 21.20 & 32.74 \\
        \midrule
        42 & C3 & 77.86 & 23.56 & 12.97 \\
        43 & C3 & 76.31 & 25.60 & 14.82 \\
        44 & C3 & 81.22 & 19.17 & 13.62 \\
        \midrule
        42 & C4 & 81.28 & 19.66 & 14.01 \\
        44 & C4 & 80.23 & 20.54 & 14.78 \\
        45 & C4 & 80.03 & 21.04 & 12.11 \\
 
        \bottomrule
    \end{tabular}
    \endgroup
\end{table}

\noindent Fashion-MNIST behaves similarly to MNIST but with lower absolute accuracy. The key observation is that scheduler-heavy settings (C3/C4) cut spike activity by up to $63\%$ while holding accuracy within $\pm$2 percentage points of the replay baseline, confirming that the proposed controller generalizes to supplementary datasets.

\FloatBarrier

\section{Contextual Comparison to Recent Spiking Continual Learning}
\label{app:prior_work_context}

This appendix summarizes representative reported results from recent spiking continual-learning works. Table~\ref{tab:prior_work_context} provides context but is not directly comparable to our main results (Table~\ref{tab:main_results}) because the datasets, task definitions, and continual-learning settings differ (e.g., permuted-MNIST domain-incremental protocols vs.\ event-based vision class-incremental protocols). We therefore use this table to clarify scope rather than claim superiority across mismatched benchmarks.

\begin{table}[!ht]
\centering
\caption{Protocol-aware context from recent spiking continual-learning works (reported by the respective papers).}
\label{tab:prior_work_context}
\begingroup
\small
\setlength{\tabcolsep}{3pt}
\renewcommand{\arraystretch}{1.15}
\begin{tabularx}{\textwidth}{p{2.25cm} p{2.55cm} p{3.15cm} X}
\toprule
\textbf{Work} & \textbf{Setting / benchmark} & \textbf{Reported metric(s)} & \textbf{Notes on comparability to this paper} \\
\midrule
Xiao et al.\ (HLOP) \cite{xiao2024hebbian} & Permuted-MNIST (domain-incremental) & ACC 95.15\%, BWT $-1.30\%$ (DSR+HLOP, reported) & Strong retention on permuted-MNIST under domain-incremental protocols. Not event-based; does not report spike/energy trade-offs on neuromorphic vision benchmarks. \\
\midrule
Soures et al.\ (TACOS) \cite{soures2024tacos} & Split-MNIST / Split-FMNIST (task-agnostic domain-incremental) & Mean accuracy 82.56\% (MNIST) and 93.22\% (FMNIST) & Task-agnostic domain-incremental setting with a different architecture/training regime. Not evaluated on DVS vision datasets; does not analyze accuracy--energy control via spike budgeting. \\
\midrule
Han et al.\ (SOR-SNN) \cite{han2023adaptive} & CIFAR-100 incremental (5/10/20-step variants) & Example: 10-step CIFAR-100 average accuracy 80.12\% (reported) & Targets large-scale class-incremental benchmarks with pathway reorganization and energy/memory analyses, but not native event-based neuromorphic vision datasets used here. \\
\midrule
Larionov et al.\ (CoLaNET) \cite{larionov2025continual} & Permuted-MNIST (10 tasks) & Example: average accuracy $\approx$ 92.31\%, with low forgetting (reported) & Biologically plausible local-learning SNN on permuted-MNIST. Different task construction and no cross-modality neuromorphic vision evaluation. \\
\bottomrule
\end{tabularx}
\endgroup
\end{table}

\end{document}